\documentclass{article}
\usepackage{ijcai13}
\usepackage{makeidx}  
\usepackage{times}
\usepackage{graphicx}
\usepackage{latexsym,footnote}
\usepackage{amsmath}
\usepackage[ruled,vlined,linesnumbered,norelsize]{algorithm2e}
\usepackage{mathnotation}

\usepackage[T1]{fontenc} 
\usepackage{color}
\usepackage[draft]{fixme}

\usepackage{subfigure}

\fxsetup{
nomargin,inline,index,
theme=color
}

\newcommand{\tax}{\mathit{ax}}

\newcommand{\mo}{\mathcal{O}}
\newcommand{\mt}{\mathcal{T}}
\newcommand{\ma}{\mathcal{A}}
\newcommand{\mi}{\mathcal{I}}
\newcommand{\mb}{\mathcal{B}}
\newcommand{\md}{\mathcal{D}}

\newcommand{\Tp}{\mathit{P}}
\newcommand{\Tn}{\mathit{N}}
\newcommand{\tp}{\mathit{p}}
\newcommand{\tn}{\mathit{n}}

\newcommand{\EX}{EX}

\newcommand{\mX}{{\bf{X}}}

\newcommand{\dx}[1]{{\bf D}_{#1}^P}
\newcommand{\dnx}[1]{{\bf D}_{#1}^{N}}
\newcommand{\dz}[1]{{\bf D}_{#1}^\emptyset}
 
\newcommand{\qc}{\mathit{c}_q}
\newcommand{\uc}{c}

\newcommand{\ax}{{\mathit{ax}}}

\newcommand{\RQ}{{\mathit{R}}}

\newcommand{\Xsc}{\mathit{X_{sc}}}
\newcommand{\Xalt}{\mathit{X_{alt}}}

\newcommand{\Al}{{\mathcal{M}}}

\newcommand{\Sig}{{\mathbf{S}}}
\newcommand{\minD}{{\bf{MD}}}
\newcommand{\mD}{{\bf{D}}}
\newcommand{\dt}{\mathcal{D}^*}
\newcommand{\ot}{\mathcal{O}^*}
\newcommand{\otarget}{\mathcal{O}_{t}}

\newcommand{\NHR}{{\mathit{NHR}}}
\newcommand{\N}{\sigma}

\newtheorem{definition}{Definition}

\begin{document}
\title{RIO: Minimizing User Interaction in Debugging of Knowledge Bases}
\author{Patrick Rodler \and Kostyantyn Shchekotykhin \and Philipp Fleiss \and Gerhard Friedrich \vspace{3pt} \\ Alpen-Adria Universit\"at \\ Klagenfurt, Austria \\
\textit{firstname.lastname@aau.at} }

\maketitle

\begin{abstract}
The best currently known interactive debugging systems rely upon some meta-information in terms of fault probabilities in order to improve their efficiency. However, misleading meta information might result in a dramatic decrease of the performance and its assessment is only possible a-posteriori. Consequently, as long as the actual fault is unknown, there is always some risk of suboptimal interactions.
In this work we present a reinforcement learning strategy that continuously adapts its behavior depending on the performance achieved and minimizes the risk of using low-quality meta information. 
Therefore, this method is suitable for application scenarios where reliable prior fault estimates are difficult to obtain. 
Using diverse real-world knowledge bases, we show that the proposed interactive 
query strategy is scalable, features decent reaction time, and outperforms both entropy-based and no-risk strategies on average w.r.t.~required amount of user interaction. 
\end{abstract}

\section{Introduction} 
Efficient debugging is a prerequisite for successful evolution, maintenance and application of knowledge-based systems. In a standard application scenario a debugger deals with a faulty knowledge base (KB) $\mo$ which fails to meet predefined quality criteria $\RQ$ such as consistency.
The task of debugging aims at modifying $\mo$ in that a (subset-)minimal set of axioms $\md\subseteq\mo$, termed diagnosis, is deleted in order to restore compliance of the KB with $\RQ$, whereas
a set of axioms $\EX_\md$ is inserted to $\mo$ to preserve designated entailments which might have been broken by the removal of $\md$. 
Usually, a large number of competing diagnoses exist for a faulty $\mo$. Without additional information, there is no means to decide which $\md$ to prefer. In many practical scenarios, however, there is some kind of meta information available, for example in terms of (1) logs of prior debugging sessions, (2) common faults or fault patterns occurring in logical formulas, or (3) a subjective guess of the involved user based on their experience. 
Given such data, one can extract a-priori fault probabilities and use them to guide the search for diagnoses. For example, one could use a uniform cost strategy to find the most probable diagnosis w.r.t.\ fault probabilities, see e.g.~\cite{Kalyanpur2006a}. 
However, only in the best case, if the fault probabilities are perfectly adjusted for the particular case, this will lead the search to the desired diagnosis the deletion of which enables to formulate a KB compliant with the requirements defined by the user.

Interactive debugging systems such as~\cite{jws12,SiddiqiH11} tackle this issue by letting an oracle  take action during the debugging session by answering queries.
In case of KBs a debugger asks about entailments and non-entailments of the desired $\otarget$, called test cases~\cite{jws12}. These pose constraints to the validity of diagnoses and thus help to sort out incompliant diagnoses and update the probabilities of remaining ones step-by-step. 
However, often a debugger can find many alternative queries for a set of diagnoses. Selection of the ``best'' query, an answer to which allows to obtain maximum information, is very important since it  affects the total number of queries required to localize the fault.
In their seminal work \cite{dekleer1987} proposed two query selection strategies: split-in-half and entropy-based.
%
The latter strategy can make optimal profit from exploiting properly adjusted initial fault probabilities, whereas it can completely fail in the case of weak prior information. The split-in-half manifests constant behavior independently of the probabilities given, but lacks the ability to leverage appropriate fault information. Selection of the best strategy is problematic, since one has to decide about the quality of the prior fault probabilities without knowing the desired solution. 
Our evaluation shows that selection of an inappropriate strategy can result in a substantial increase of more than $2000\%$ w.r.t.\ number of queries.

The contribution of this paper is a new RIsk Optimization reinforcement learning method (RIO). Compared to existing strategies RIO allows to minimize user interaction in the average case 
for \emph{any} quality of meta information. By virtue of its learning capability, our approach is optimally suited for debugging of KBs where only vague or no meta information is available. 
Moreover, RIO uses the acquired information to adapt its learning strategy. On the one hand, our method takes advantage of the given meta information as long as good performance is achieved. On the other hand, it gradually gets more independent of meta information if suboptimal behavior is measured. 
%
Experiments on two datasets of faulty ontologies show the feasibility, efficiency and scalability of RIO. The evaluation will indicate that, on average, RIO is the best choice of strategy for both good and bad meta information with savings as to user interaction of up to 80\%.

Technical preliminaries are provided in Section~\ref{sec:basics}. Section~\ref{sec:theory} explains the suggested approach and gives implementation details. Evaluation results are described in Section~\ref{sec:eval}. Section \ref{sec:conclusion} concludes.

\section{Preliminaries} \label{sec:basics}
In order to make the paper self-contained we provide a short introduction to description logic (DL), which is a knowledge representation and reasoning system (KRS) used in the paper. Of course, the approach suggested in this work is not limited to DL and can be applied to any KRS for which there is a sound and complete reasoning method and the entailment relation is extensive, monotone and idempotent. 


\emph{Description logic}~\cite{DLHandbook} is a family of knowledge representation languages with a formal logic-based semantics that are designed to represent knowledge about a domain in form of concept descriptions. 
The syntax of a language $\mathcal{L}$ is defined by its signature (vocabulary) and a set of constructors. A signature in this case corresponds to a (disjoint) union of sets $N_C$, $N_R$ and $N_I$, where $N_C$ contains all concept names (unary predicates), $N_R$ comprises all role names (binary predicates) and $N_I$ is a set of individuals (constants). Each concept and role description can be either atomic or complex. The latter ones are composed using constructors defined in the particular language $\mathcal{L}$. 
A typical set of DL constructors includes conjunction $A \sqcap B$, disjunction $A \sqcup B$, negation $\lnot A$, existential $\exists r.A$ and value $\forall r.A$ restrictions, where $A,B \in N_C$ and $r \in N_R$. 

A DL ontology $\mo$ is defined as a tuple $(\mt, \ma)$, where $\mt$ (TBox) is a set of terminological axioms and $\ma$ (ABox) a set of assertional axioms. Each TBox axiom is expressed by a general concept inclusion $A \sqsubseteq C$, a form of logical implication, or by a definition $A \equiv C$, a kind of logical equivalence, where $C$ is an atomic or complex concept. 
ABox axioms are used to assert properties of individuals in terms of the vocabulary defined in TBox, e.g.\ concept $A(x)$ or role $r(x,y)$ assertions, where $x,y \in N_I$.

The semantics of DLs is given in terms of interpretations $\mi =(\Delta^\mi, \cdot^\mi)$ consisting of a non-empty domain $\Delta^\mi$ and a function $\cdot^\mi$ that maps each concept to a subset of $\Delta^\mi$, each role to a subset of $\Delta^\mi \times \Delta^\mi$ and each individual to some value in $\Delta^\mi$. An interpretation $\mi$ is a model of $\mo$ iff it satisfies all TBox and ABox axioms. $\mo$ is unsatisfiable iff it has no model. A concept $A$ (role $r$) is satisfiable w.r.t\ $\mo$ iff there is a model $\mi$ of $\mo$ with $A^\mi \neq \emptyset$ ($r^\mi \neq \emptyset$). A TBox is incoherent iff there exists an unsatisfiable concept or role.

Usually description logic systems provide sound and complete reasoning services to their users. In addition to verification of coherence and consistency of $\mo$, the reasoners also perform classification and realization. Classification is a subsumption algorithm that determines most specific (general) concepts that subsume (are subsumed by) a certain concept. Realization computes for each individual $x$ a set of most specific concepts $\setof{C_1,\dots,C_n}$ such that $\mo\models C_i(x)$ for all $i=1,\dots,n$. 
Note, when we speak of entailments below, we address (only) the output computed by the classification and realization services of a DL-reasoner.

\emph{Ontology debugging}, given an ontology $\mo$, aims at approximating the so-called \emph{target ontology} $\otarget$ by $\ot$, where 
$\otarget$ is some correct and complete ontology that satisfies \emph{all} requirements to the knowledge-based application it is used for.
$\ot$ must satisfy all \emph{explicitly} stated requirements and is thus termed \emph{complying ontology}. It results from modifications to $\mo$ in terms of (1)~deleting axioms $\md$ and (2)~inserting axioms $EX_\md$. 
We call $\md=\mo\setminus\ot$ a diagnosis.
%
\begin{definition}[Complying Ontology, Diagnosis Problem] Let $\mo$ be an ontology, $\mb$ a background KB, $\RQ$ a set of requirements
to $\mo$, $\Tp$ and $\Tn$ respectively a set of positive and negative test cases, where each test case $p\in\Tp$ and $n\in\Tn$ is a set of axioms.
Then an ontology $\ot$ is called \emph{complying ontology} iff all the following conditions hold:
\vspace{-5pt}
\begin{eqnarray}
		 \forall \, r  \in \RQ&:& \;\ot \cup \mb \,\text{ fulfills }\, r  \\
		 \forall \,\tp \in \Tp&:& \;\ot \cup \mb \,\models\, \tp				\\
		 \forall \,\tn \in \Tn&:& \;\ot \cup \mb \,\not\models\, \tn 
\end{eqnarray}
\vspace{-14pt}\\
The tuple $\langle\mo,\mb,\Tp,\Tn\rangle_\RQ$ defines a \emph{diagnosis problem instance (DPI)}. 
\end{definition} 
Often $\RQ:=\{\text{coherence},\text{consistency}\}$ is assumed. 

\begin{definition}[Diagnosis]\label{def:diagnosis} 
$\md \subseteq \mo$ is called \emph{diagnosis} for a DPI $\langle\mo,\mb,\Tp,\Tn\rangle_\RQ$ iff there is a set of axioms $EX_\md$ such that $(\mo\setminus\md)\cup EX_\md$ is a complying ontology.
A diagnosis $\md$ assumes that all $ax_i \in \md$ are faulty and all $ax_j \in \mo\setminus\md$ are correct. A diagnosis $\md$ is \emph{minimal} iff there is no $\md' \subset \md$ s.t.\ $\md'$ is a diagnosis. $\minD$ denotes the set of minimal diagnoses of a DPI.
\end{definition}
Note that $\minD$ is usually used to approximate the set of all diagnoses of a DPI.
The identification of $EX_\md$, accomplished e.g.\ by some learning approach, is a crucial part of the ontology repair process. However, the complete formulation of $EX_\md$ is outside the scope of this work where we focus on computing diagnoses. As suggested in~\cite{jws12}, we approximate $EX_\md$ by the set $\bigcup_{\tp \in \Tp} \tp$. Given a DPI $\langle\mo,\mb,\Tp,\Tn\rangle_\RQ$, if the set of axioms $\mo \cup \bigcup_{\tp \in \Tp} \tp$ is not a complying ontology then 
there is no diagnosis $\md = \emptyset$, i.e.\ some axioms in $\mo$ must be modified.  

\noindent\textbf{Example 1:} 
Consider 
$\mo:=\mo_1\cup\mo_2\cup\Al_{12}$ with TBox $\mt$:
\begin{center}
\footnotesize
\begin{tabular}{crl}
$\mo_1$ & $\tax_1:$ &  $PhD \sqsubseteq Researcher$ \\ 
        & $\tax_2:$ &  $Researcher \sqsubseteq DeptEmployee$ \vspace{1pt}\\ 
        \hline\vspace{-8pt}\\
$\mo_2$ & $\tax_3:$ &  $PhDStudent \sqsubseteq Student$  \\ 
        & $\tax_4:$ &  $Student \sqsubseteq \lnot DeptMember$ \vspace{1pt}\\ 
        \hline\vspace{-8pt}\\
$\Al_{12}$ & $\tax_5:$ &  $PhDStudent \sqsubseteq PhD$ \\
        & $\tax_6:$ &  $DeptEmployee \sqsubseteq DeptMember$
\end{tabular}
\end{center}
and ABox $\ma=\setof{PhDStudent(s)}$, where $\Al_{12}$ is an automatically generated set of semantic links between $\mo_1$ and $\mo_2$. The given ontology $\mo$ is inconsistent since it describes $s$ as both a department member and not. 
Let the DPI be defined as $\tuple{\mt,\ma,\emptyset,\emptyset}_{\setof{\text{coherence}}}$, where $\ma$ is correct and thus added to the background theory
and both sets $\Tp$ and $\Tn$ are empty. 
For this DPI    
$\minD=\{\md_1 : [\tax_1], \md_2 : [\tax_2], \md_3 : [\tax_3], \md_4 : [\tax_4], \md_5 : [\tax_5],\md_6 : [\tax_6]\}$. 
To compute $\minD$ we employ a combination of HS-Tree~\cite{Reiter87} and QuickXPlain~\cite{junker04} algorithms as suggested by \cite{friedrich2005gdm}.
%

\emph{Interactive ontology debugging} iteratively incorporates a user's knowledge about $\otarget$, thereby differentiating between diagnoses in $\minD$. The overall procedure is as follows: (1)~Compute a set of at most $n$ leading diagnoses $\mD \subseteq \minD$ that serve as an approximation of all minimal diagnoses $\minD$. 
Restricting the computation of 
$\minD$ to a predefined number $n$ helps to overcome exponential explosion of HS-Tree.
Preference criteria such as most probable or minimum cardinality diagnoses are used to specify 
$\mD$ within $\minD$.
(2)~Exploit $\mD$ to compute/select a query which is posed to the user. (3)~Incorporate the user's answer to 
prune
the search space for diagnoses. Go to (1) until a predefined stop criterion is met by a $\dt\in\mD$, e.g.~$\dt$ has overwhelming probability. 
We call the priorly unknown diagnosis that will meet the stop criterion \emph{target diagnosis} $\dt$.
As a means for interaction with the user we utilize the notion of a query which means asking the user $(\otarget \models X_j ?)$, i.e.~to classify whether a given set of axioms $X_j$ should be entailed (assigned to $\Tp$) or not entailed (assigned to $\Tn$) by $\otarget$.
The theoretical foundation for the application of queries is the fact  
that $\mo\setminus\md_i$ and $\mo\setminus\md_j$ for $\md_i \neq \md_j \in \mD$\, entail different sets of axioms. 
%
\begin{definition}[Query, Partition]
Let $\mo^{*}_i := (\mo \setminus \md_i) \cup \mb \cup (\bigcup_{\tp\in\Tp} \tp)$ where $\md_i\in\mD$. A set of axioms $X_j$ is called a \emph{query} iff $\dx{j}:=\setof{\md_i \in \mD\,|\,\mo^{*}_i \models X_j}\neq \emptyset$ and $\dnx{j}:=\setof{\md_i \in \mD\,|\,\mo^{*}_i \models \lnot X_j}\neq \emptyset$.
The 
\emph{partition} of query $X_j$ is denoted by $\langle \dx{j}, \dnx{j}, \dz{j} \rangle$ where $\dz{j} = \mD \setminus (\dx{j} \cup \dnx{j})$.
$\mX_\mD$ terms the set of all queries and associated partitions w.r.t. $\mD$.
\end{definition}
%
The (complete) set 
$\mX_\mD$ can be generated as shown in Algorithm~\ref{algo_query_gen}. In each iteration, given a set of diagnoses $\dx{k} \subset \mD$, common entailments $X_k:=\setof{e\,|\,\forall\md_i\in\dx{k}:\mo^*_i\models e}$ are computed (function \textsc{getEntailments}) and used to classify the remaining diagnoses in $\mD\setminus\dx{k}$ to obtain the partition $\langle\dx{k},\dnx{k},\dz{k}\rangle$ associated with~$X_k$. 
Then, together with its partition, $X_k$ is added to 
$\mX_\mD$. 
The function \textsc{inconsist}($arg$) returns $true$ if $arg$ is inconsistent or incoherent. 
%

Let the answering of queries by a user be modeled as function $u: \mX_\mD \rightarrow \{\textit{t},\textit{f}\}$. If $u_j := u(X_j) = \textit{t}$, then $\Tp \leftarrow \Tp \cup \setof{X_j}$ and $\mD \leftarrow \mD\setminus\dnx{j}$. Otherwise, $\Tn \leftarrow \Tn \cup \setof{X_j}$ and $\mD \leftarrow \mD\setminus\dx{j}$.
Prospectively, according to Definition \ref{def:diagnosis}, only those diagnoses are considered in the set $\mD$ that comply with the new DPI obtained by the 
addition
of a test case.
%
This allows us to formalize the problem we address in this work:

\vspace{3pt}
\noindent\textbf{Problem Definition (Diagnosis Discrimination)}
\emph{\textbf{Given}~$\mD$ w.r.t. $\langle\mo,\mb,\Tp,\Tn\rangle_\RQ$, a stop criterion $stop:\mD\rightarrow\{t,f\}$ and a user~$u$, \textbf{find} a next query $X_j\in\mX_\mD$ such that 
(1)~$(X_j,\dots,X_q)$ is a sequence of minimal length and (2)~after $X\in\setof{X_j,\dots,X_q}$ are added to $\Tp$ and $\Tn$ according to $\setof{u_j,\dots,u_q}$, there
exists a $\dt \in \mD$ such that $stop(\dt) = t$. 
}
\begin{algorithm}[tb]
\scriptsize
\KwIn{DPI $\tuple{\mo,\mb,\Tp,\Tn}_\RQ$, set of corresponding diagnoses $\mD$}
\KwOut{a set of queries and associated partitions $\mX_\mD$} 
\SetKwFunction{getEnts}{getEntailments}
\SetKwFunction{inconsist}{inconsist}
\ForEach{$\dx{k} \subset \mD$}{
      $X_k \leftarrow \getEnts(\mo, \mb, \Tp, \dx{k})$\;
			\If {$X_k \neq \emptyset$} {
				\ForEach {$\md_r \in \mD\setminus\dx{k}$}{ 
					\lIf { $\mo^{*}_r \,\models X_k$}{$\dx{k} \leftarrow \dx{k} \cup \left\{\md_r\right\}$}\;
					\lElseIf	
					{$\inconsist(\mo^{*}_r \cup X_k)$}
					{$\dnx{k} \leftarrow \dnx{k} \cup \left\{\md_r\right\}$}\;
					\lElse {$\dz{k} \leftarrow \dz{k} \cup \left\{\md_r\right\}$}\;
				} 
			$\mX_\mD \leftarrow \mX_\mD \cup \tuple{X_k, \tuple{\dx{k}, \dnx{k}, \dz{k}}}$
    } 
}
\Return $\mX_\mD$\;
\caption{\small Query Generation \normalsize} \label{algo_query_gen}
\normalsize
\end{algorithm}
%

Two strategies for selecting the ``best'' next query have been proposed \cite{dekleer1987} and adapted to debugging of KBs by~\cite{jws12}. 
\noindent\textbf{Split-in-half strategy (SPL)}, selects the query $X_j\in\mX_\mD$ which minimizes the 
scoring function $sc_{split}(X_j) := \left| |\dx{j}| - |\dnx{j}| \right| + |\dz{j}|$.
So, SPL prefers queries which eliminate half of the diagnoses independently of the query outcome.
%
%
\noindent\textbf{Entropy-based strategy (ENT)} 
uses information about prior probabilities $p_t$ for the user to make a mistake when using a syntactical construct of type $t \in \mathit{CT}(\mathcal{L})$, where $\mathit{CT}(\mathcal{L})$ is the set of constructors available in the used knowledge representation language $\mathcal{L}$, e.g.~$\setof{\forall, \exists, \sqsubseteq, \neg, \sqcup, \sqcap} \subset \mathit{CT}(\text{OWL})$~\cite{Grau2008a}. These fault probabilities $p_t$ are assumed to be independent and used to calculate fault probabilities of axioms $ax_k$ as 
$p(\ax_k) = 1 - \prod_{t \in \mathit{CT}} (1-p_t)^{n(t)}$
 where $n(t)$ is the number of occurrences of construct type $t$ in $ax_k$.
The probabilities of axioms can in turn be used to determine fault probabilities of diagnoses $\md_i \in \mD$ as 
\vspace{-1pt}
\begin{equation}
\label{eq:prob_diagnosis}
p(\md_i) = \prod_{\ax_r \in \md_i} p(\ax_r) \prod_{\ax_s \in \mo\setminus\md_i} (1-p(\ax_s))
\end{equation}
\vspace{-1pt}\\
ENT selects the query $X_j\in\mX_\mD$ with highest expected information gain, i.e. which minimizes $sc_{ent}(X_j)$ defined as:
\vspace{-1pt}
\begin{equation*}
\sum_{a\in\{t,f\}} p(u_j = a) \sum_{\md_k \in {\bf D}} -p(\md_k | u_j = a) \log_2 p(\md_k | u_j = a)
\vspace{2pt}
\end{equation*}
where 
$p(u_j=t) = \sum_{\md_r \in \dx{j}} p(\md_r) + \frac{1}{2} p(\dz{j})$
, $p(\dz{j})=\sum_{\md_r \in \dz{j}} p(\md_r)$
and $p(u_j=f)=1-p(u_j=t)$.
%
The answer $u_j=a$ 
is used to update probabilities $p(\md_k)$ according to the Bayesian formula, yielding $p(\md_k | u_j = a)$.
%
%
The result of the evaluation in~\cite{jws12} shows that ENT reveals better performance than SPL in most of the cases. However, SPL proved to be the best strategy in situations when 
misleading prior information is provided, i.e.~the target diagnosis $\dt$ has low probability. So, one can regard ENT as a high risk strategy with high potential to perform well, depending on the priorly unknown quality of the given fault information. SPL, in contrast, can be seen as a no-risk strategy without any potential to leverage good meta information.
Therefore, selection of the proper combination of prior probabilities $\setof{p_t\,|\,t\in\mathit{CT}(\mathcal{L})}$ and query selection strategy is crucial for successful 
diagnosis discrimination
and minimization of user interaction.

\section{Risk Optimization for Query Selection}
\label{sec:theory} 
The proposed Risk Optimization Algorithm (RIO) extends ENT strategy with a dynamic learning procedure that learns by reinforcement how to select optimal queries. The behavior is determined by the achieved performance in terms of diagnosis elimination rate. Good performance means similar behavior to ENT, whereas aggravation of performance leads to a gradual neglect of the given meta information. Like ENT, RIO continually improves the prior fault probabilities based on new knowledge obtained through queries to a user.

RIO learns a ``cautiousness'' parameter $\uc$ whose admissible values are captured by the user-defined interval $[\underline{\uc},\overline{\uc}]$. The relationship between $\uc$ and queries is as follows:
%
\begin{definition}[Cautiousness of a Query]\label{def:query_cautiousness}
We define the \emph{cautiousness} $\qc(X_i)$ of a query $X_i$ as follows:
\vspace{-7pt}
\begin{equation*}\label{eq:query_cautiousness}
\qc(X_i) := \frac{ \min\left\{|\dx{i}| ,|\dnx{i}| \right\} }{|\mD|} \in \left[0,\frac{\left\lfloor \frac{|\mD|}{2}\right\rfloor}{|\mD|}\right] =: [\underline{\qc},\overline{\qc}]
\vspace{-3pt}
\end{equation*}
A query $X_i$ is called \emph{braver} than query $X_j$ iff $\qc(X_i) < \qc(X_j)$. Otherwise $X_i$ is called \emph{more cautious} than $X_j$.
A query with 
maximum cautiousness $\overline{\qc}$ is called \emph{no-risk query}.
\end{definition}
\begin{definition}[Elimination Rate]\label{def:elimination_rate}
Given a query $X_i$ and the corresponding answer $u_i \in \{t,f\}$, the \emph{elimination rate} $e(X_i,u_i)=\frac{|\dnx{i}|}{|\mD|}$ if $u_i = t$ and $e(X_i,u_i)=\frac{|\dx{i}|}{|\mD|}$ if $u_i = f$. 
The answer $u_i$ to a query $X_i$ is called \emph{favorable} iff it maximizes the elimination rate $e(X_i,u_i)$. Otherwise $u_i$ is called \emph{unfavorable}. The minimal or \emph{worst case elimination rate} $\min_{u_i \in \{t,f\}}(e(X_i,u_i))$ of $X_i$ is denoted by $e_{wc}(X_i)$.
\end{definition}
So, the cautiousness $\qc(X_i)$ of a query $X_i$ is exactly the worst case elimination rate, i.e. $\qc(X_i) = e_{wc}(X_i) = e(X_i,u_i)$ given that $u_i$ is the unfavorable query result.
Intuitively, parameter $\uc$ characterizes the minimum proportion of diagnoses in $\mD$ which should be eliminated by the successive query.
%
\begin{definition}[High-Risk Query]\label{def:high-risk_query}
Given a query $X_i$ and cautiousness $\uc$, $X_i$ is called a \emph{high-risk query} iff $\qc(X_i) < \uc$, i.e. the cautiousness of the query is lower than the algorithm's current cautiousness value $c$.
Otherwise, $X_i$ is called \emph{non-high-risk query}. By $\NHR_c(\mX_\mD)\subseteq \mX_\mD$ we denote the set of all non-high-risk queries w.r.t. $c$. For given cautiousness $c$, the set of all queries $\mX_\mD$ can be partitioned in high-risk queries and non-high-risk queries.
\end{definition}

\label{ex:high-risk_query}
\noindent\textbf{Example 2 (cont. Example 1):} 
Let the user specify $\uc := 0.3$ for the set $\mD$ with $|\mD|=6$. 
Given these settings, $X_1:=\{DeptEmployee(s),Student(s)\}$ is a non-high-risk query since its partition $\langle \dx{1}, \dnx{1}, \dz{1} \rangle = \langle \setof{\md_4,\md_6}, \setof{\md_1,\md_2,\md_3,\md_5}, \emptyset\rangle$ and thus its cautiousness $\qc(X_1) = 2/6 \geq 0.3 = \uc$. The query $X_2 :=\{PhD(s)\}$ with partition $\langle \setof{\md_1,\md_2,\md_3,\md_4,\md_6}, \setof{\md_5}, \emptyset\rangle$ is a high-risk query because $\qc(X_2) = 1/6 < 0.3 = \uc$ and $X_3:=\{Researcher(s),Student(s)\}$ with $\langle \setof{\md_2,\md_4,\md_6}, \setof{\md_1,\md_3,\md_5}, \emptyset\rangle$ is a no-risk query due to $\qc(X_3) = 3/6 = \overline{\qc}$.

Given a user's answer $u_s$ to a query $X_s$, the cautiousness $\uc$ is updated depending on the elimination rate $e(X_s,u_s)$ by $\uc \leftarrow \uc + \uc_{adj}$
%
%
where 
the cautiousness adjustment factor $\uc_{adj} :=\; 2\,(\overline{\uc} - \underline{\uc})\mathit{adj}$.
%
%
The scaling factor $2 \, (\overline{\uc} - \underline{\uc})$ 
regulates the extent of the cautiousness adjustment depending on the interval length $\overline{\uc} - \underline{\uc}$. 
More crucial is the factor $\mathit{adj}$ that indicates the sign and magnitude of the cautiousness adjustment.  
\vspace{-5pt}
\begin{equation*}
\mathit{adj} :=  \frac{\left\lfloor \frac{|\mD|}{2}-\varepsilon\right\rfloor}{|\mD|} - e(X_s,u_s)
\vspace{-5pt}
\end{equation*}
where $\varepsilon \in (0,\frac{1}{2})$ is a constant which prevents the algorithm from getting stuck in a no-risk strategy for even $|\mD|$. E.g., given $\uc=0.5$ and $\varepsilon=0$, the elimination rate of a no-risk query $e(X_s,u_s) = \frac{1}{2}$ resulting always in $adj=0$. 
The value of $\varepsilon$ can be set to an arbitrary real number, e.g. $\varepsilon := \frac{1}{4}$. 
%
%
If $\uc + \uc_{adj}$
is outside the user-defined cautiousness interval $[\underline{\uc},\overline{\uc}]$, it is set to $\underline{\uc}$ if $\uc < \underline{\uc}$ and to $\overline{\uc}$ if $\uc > \overline{\uc}$. Positive $\uc_{adj}$ is a penalty telling the algorithm to get more cautious, whereas negative $\uc_{adj}$ is a bonus resulting in a braver behavior of the algorithm.
Note, for the user-defined interval $[\underline{\uc},\overline{\uc}] \subseteq [\underline{\qc},\overline{\qc}]$ must hold. 
$\underline{\uc} - \underline{\qc}$ and $\overline{\qc} - \overline{\uc}$ represent the minimal desired difference in performance to a high-risk (ENT) and no-risk (SPL) query selection, respectively.
By expressing trust (disbelief) in the prior fault probabilities through specification of lower (higher) values for $\underline{\uc}$ and/or $\overline{\uc}$, the user can take influence on the behavior of RIO.
 
\noindent\textbf{Example 3 (cont. Example 1):}  Assume $p(\tax_i) := 0.001$ for $ax_{i(i=1,\dots,4)}$ and $p(\tax_5):=0.1, p(\tax_6):=0.15$ and the user 
rather disbelieves these fault probabilities and thus
sets $\uc=0.4$, $\underline{\uc}=0$ and $\overline{\uc}=0.5$. In this case RIO selects a no-risk query $X_3$ just as SPL. Given 
$u_3=t$ and $|\mD|=6$, the algorithm computes the elimination rate $e(X_3,t)=0.5$ and adjusts the cautiousness by $\uc_{adj}=-0.17$ which yields $\uc=0.23$. This allows RIO to select a higher-risk query in the next iteration, whereupon the
target diagnosis $\dt=\md_2$ is found after asking three queries. In the same situation, ENT (starting with high-risk query $X_1$) would require four queries.

\begin{algorithm}[b]
\scriptsize
\KwIn{diagnosis problem instance $\langle\mo,\mb,\Tp,\Tn\rangle_\RQ$, fault probabilities of diagnoses $DP$, cautiousness $C=(\uc,\underline{\uc},\overline{\uc})$, number of leading diagnoses $n$ to be considered, acceptance threshold $\sigma$
}
\KwOut{a diagnosis $\md$} 

\SetKwFunction{getMinScQ}{getMinScoreQuery}
\SetKwFunction{getScore}{eliminationRate}
\SetKwFunction{aboveThresh}{aboveThreshold}
\SetKwFunction{mostProbDiag}{mostProbableDiag}
\SetKwFunction{getPercent}{getQueryCautiousness}
\SetKwFunction{getAlpha}{getMinAlpha}
\SetKwFunction{getBestQ}{getAlternativeQuery}
\SetKwFunction{performAdapt}{performAdaptation}
\SetKwFunction{getAnswer}{getAnswer}
\SetKwFunction{updateProbs}{updateProbablities}
\SetKwFunction{updateRisk}{updateCautiousness}
\SetKwFunction{getDiagnoses}{getDiagnoses}
\SetKwFunction{genQs}{generateQueries}
\SetKwFunction{compPriors}{getProbabilities}
$\Tp \leftarrow \emptyset$;
$\Tn \leftarrow \emptyset$;
$\mD \leftarrow \emptyset$\;
\Repeat{$(\aboveThresh(DP,\sigma) \lor \getScore(X_s) = 0$)}{
        $\mD \leftarrow \getDiagnoses(\mD, n, \mo, \mb, \Tp, \Tn)$\;
				$DP \leftarrow \compPriors(DP,\mD, \Tp, \Tn)$\;
				$\mX \leftarrow \genQs(\mo, \mb, \Tp, \mD)$\;
				$X_s \leftarrow \getMinScQ(DP,\mX)$\;    
				\lIf {$\getPercent(X_s,\mD) < \uc$}{        
						$X_s \leftarrow \getBestQ(\uc,\mX, DP, \mD)$\;
				}  
				\lIf {$ \getAnswer(X_s) = \textit{yes}$}{        
						$\Tp \leftarrow \Tp \cup \{X_s\}$\;  
				}
				\lElse{
						$\Tn \leftarrow \Tn \cup \{X_s\}$\;
				} 
				$c \leftarrow \updateRisk(\mD,\Tp,\Tn,X_s,\uc,\underline{\uc},\overline{\uc})$\;
}
\Return $\mostProbDiag(\mD, DP)$\;
\caption{Risk Optimization Algorithm (RIO)} \label{algo_main}
\normalsize
\end{algorithm}

RIO, described in Algorithm \ref{algo_main}, starts with the computation of minimal diagnoses. \textsc{getDiagnoses} function implements a combination of HS-Tree and QuickXPlain algorithms. Using uniform-cost search, the algorithm extends the set of leading diagnoses $\mD$ with a maximum number of most probable minimal diagnoses such that $|\mD| \leq n$. 

Then the \textsc{getProbabilities} function calculates the fault probabilities $p(\md_i)$ for each diagnosis $\md_i$ of the set of leading diagnoses $\mD$ using formula~(\ref{eq:prob_diagnosis}). Next it adjusts the probabilities as per the Bayesian theorem taking into account all previous query answers which are stored in $\Tp$ and $\Tn$. Finally, the resulting probabilities $p_{adj}(\md_i)$ are normalized.
Based on the set of leading diagnoses $\mD$, \textsc{generateQueries} generates queries according to Algorithm \ref{algo_query_gen}.
	\textsc{getMinScoreQuery} determines the best query $\Xsc \in \mX_\mD$ according to $sc_{ent}$: $\Xsc = \argmin_{X_k \in \mX_\mD}(sc_{ent}(X_k))$.
If $\Xsc$ is a non-high-risk query, i.e. $\uc \leq \qc(\Xsc)$ (determined by \textsc{getQueryCautiousness}), $\Xsc$ is selected. In this case, $\Xsc$ is the query with best information gain 
in $\mX_\mD$ and moreover guarantees the required elimination rate specified by $\uc$.
	
\label{step_alternative} Otherwise, \textsc{getAlternativeQuery} selects the query $\Xalt \in \mX_\mD$\; $(\Xalt \neq \Xsc)$ which has minimal score $sc_{ent}$ among all least cautious non-high-risk queries $L_c$. That is, $\Xalt = \argmin_{X_k \in \mathit{L}_c}(sc_{ent}(X_k)) $
where $\mathit{L_c} := \{X_r \in \NHR_c(\mX_\mD) \;|\; \forall X_t \in \NHR_c(\mX):\, \qc(X_r) \leq \qc(X_t)\}$. If there is no such query $\Xalt\in\mX_\mD$, then $\Xsc$ is selected.
	
Given the 
user's answer $u_s$, the selected query $X_s \in \setof{\Xsc,\Xalt}$ is added to $\Tp$ or $\Tn$ accordingly. 
In the last step of the main loop the algorithm updates the cautiousness value~$\uc$ (function \textsc{updateCautiousness}) as described above.

Before the next query selection iteration starts, a stop condition test is performed. The algorithm evaluates whether the most probable diagnosis is at least $\sigma\%$ more likely than the second most probable diagnosis (\textsc{aboveThreshold}) or none of the leading diagnoses has been eliminated by the previous query, i.e.\textsc{getEliminationRate} returns zero for $X_s$. 
If a stop condition is met, the presently most likely diagnosis is returned (\textsc{mostProbableDiag}).

\section{Evaluation} \label{sec:eval}
\textbf{Goals.}
This evaluation should demonstrate that (1) there is a significant discrepancy between SPL and ENT concerning number of queries where the winner depends on the quality of meta information, (2) RIO
exhibits superior average behavior compared to ENT and SPL w.r.t. the amount of user interaction required, irrespective of the quality of specified fault information, (3) RIO scales well and (4) its reaction time is well suited for an interactive debugging approach.

\vspace{2pt}
\noindent\textbf{Provenance of test data.} As data source for the evaluation we used faulty real-world ontologies produced by automatic ontology matching systems (OMSs) (cf.\ Example 1). 
%
\begin{definition}[Ontology matching]\label{def:align}~\cite{Shvaiko2012}
Let $Q(\mo)\subseteq \Sig(\mo)$ denote the set of matchable elements in an ontology $\mo$, where $\Sig(\mo)$ denotes the signature of $\mo$. An ontology matching operation determines an \emph{alignment} $\Al_{ij}$, which is a set of correspondences between matched ontologies $\mo_i$ and $\mo_j$. Each \emph{correspondence} is a 4-tuple $\tuple{x_i,x_j,r,v}$, such that $x_i \in Q(\mo_i)$, $x_j \in Q(\mo_j)$, $r$ is a semantic relation and $v \in [0,1]$ is a confidence value. We call $\mo_{i\Al j} := \mo_i \cup \N(\Al_{ij}) \cup \mo_j$ the \emph{aligned ontology} for $\mo_i$ and $\mo_j$ where $\N$ maps each correspondence to an axiom.
\end{definition}
Let in the following $Q(\mo)$ be the restriction to atomic concepts and roles in $\Sig(\mo)$, $r \in \setof{\sqsubseteq, \sqsupseteq, \equiv}$ and $\N$ the natural alignment semantics~\cite{MeilickeStuck2009} that maps correspondences one-to-one to axioms of the form $x_i~r~x_j$.
%
We 
evaluate RIO using aligned ontologies 
by the following reasons: (1)~Alignments often cause inconsistency/incoherence of ontologies. 
(2)~The (fault) structure of different ontologies obtained through matching generally varies due to different authors and matching systems involved. (3)~For the same reasons, it is hard to estimate the quality of 
fault probabilities, 
i.e. it is unclear which existing query selection strategy to choose for best performance. (4)~Availability of correct reference alignments. 

\vspace{2pt}
\noindent\textbf{Test datasets.} We used two datasets D1 and D2: Each faulty aligned ontology $\mo_{i\Al j}$ in D1 is the result of applying one of four
OMSs to a set of six independently created ontologies 
in the domain of conference organization. For a given pair of ontologies $\mo_i\neq\mo_j$, each system produced an alignment $\Al_{ij}$. The average size of $\mo_{i\Al j}$ per matching system was between $312$ and $377$ axioms. D1 is a superset of the dataset used in \cite{Stuckenschmidt2008} for which all debugging systems under evaluation manifested correctness or scalability problems. D2, used to assess the scalability of RIO, is the set of ontologies
from the ANATOMY track in the Ontology Alignment Evaluation Initiative\footnote{http://oaei.ontologymatching.org} (OAEI) 2011.5 \cite{Shvaiko2012}, which comprises two input ontologies $\mo_1$ (11545 axioms) and $\mo_2$ (4838 axioms). The size of the aligned ontologies generated by results of seven different OMSs was between 17530 and 17844 axioms. 

\vspace{0pt}
\noindent\textbf{Reference Solutions.} For dataset D1, based on a manually produced reference alignment $\mathcal{R}_{ij} \subseteq \Al_{ij}$ for ontologies $\mo_i,\mo_j$ (cf. \cite{Meilicke2008a}), we were able to fix a target diagnosis $\dt:=\N(\Al_{ij}\setminus\mathcal{R}_{ij})$ for each incoherent $\mo_{i\Al j}$. In cases where 
$\dt$
represented a non-minimal diagnosis, 
it was randomly redefined 
as 
a
minimum diagnosis
$\dt \subset \N(\Al_{ij}\setminus \mathcal{R}_{ij})$. 
In case of D2, given ontologies $\mo_1$ and $\mo_2$, 
matching output $\Al_{12}$, and the correct reference alignment $\mathcal{R}_{12}$, we fixed $\dt$ as follows: We carried out (prior to the actual experiment) a debugging session with DPI $\tuple{\N(\Al_{12}\setminus \mathcal{R}_{12}),
        \mo_1\cup\mo_2\cup \N(\Al_{12} \cap \mathcal{R}_{12}),
        \emptyset,
        \emptyset}_{\setof{\text{coherence}}}$
and randomly chose one of the identified diagnoses as $\dt$.
Note, it is common in OMS~\cite{meilicke2011} that $\dt$ can be a subset of $\md := \sigma(\Al_{ij}\setminus\mathcal{R}_{ij})$ as there is no evidence based on coherence to classify any $\tax\in\sigma(\md \setminus \dt)$ as faulty.


\begin{figure*}[ht]
\centering
\subfigure[]{
    {\fontsize{7pt}{8pt}\selectfont
    \begin{tabular}{@{\extracolsep{-4pt}} l|c|c||c|c}
																												& EXP-1		& EXP-2 	& EXP-3			& EXP-4			\\	\hline
			$q_\text{SPL} < q_\text{ENT}$ 										& 12\%	  & 37\% 		& 	0\%			&   29\%		\\
			$q_\text{ENT} < q_\text{SPL}$ 										& 81\%    &	56\% 	  & 	100\%		&		71\%		\\
			$q_\text{SPL} = q_\text{ENT}$											& 7\%     &	7\% 	  & 	0\%			&		0\%			\\
			$q_\text{RIO} < \min$															& 4\%	  	& 26\% 	  & 	29\%		&		71\%		\\
			$q_\text{RIO} \leq \min$															& 74\%		& 74\% 	  & 	100\%		&		100\%		\\	\hline 
																											
		\end{tabular}
		}
		\label{tab:best_strategies}
}
\subfigure[]{
{\fontsize{7pt}{8pt}\selectfont
\begin{tabular}{@{\extracolsep{-3pt}} l|c|c|c||c|c|c||c|c|c||c|c|c}
&  \multicolumn{3}{c||}{EXP-1}	         &  \multicolumn{3}{c||}{EXP-2}	            &  \multicolumn{3}{c||}{EXP-3}	         &  \multicolumn{3}{c}{EXP-4}                   \\ \hline
& \textit{debug} & \textit{react} & $q$  & 	\textit{debug} & \textit{react} & $q$ 	& \textit{debug} & \textit{react} & $q$  &  \textit{debug}	& \textit{react} &	 $q$	\\\hline
ENT &  1.86		 & 0.26 		      & 3.67 & 	1.42  		   & 0.20 			& 5.26	& \textbf{60.93} & 12.37		  & 5.86 & 74.46 			& 5.63 		     & 11.86	\\
SPL &  \textbf{1.43}& \textbf{0.16}& 5.70 &	\textbf{1.24}  & \textbf{0.15}   & 5.44	& 104.91	  	 & \textbf{4.79}  & 19.43& 98.65 		 	& \textbf{4.78}	 & 18.29	\\
RIO &  1.59			& 0.29 		  & \textbf{3.00}& 	1.75   & 0.25  			& \textbf{4.37}	& 62.29	 &	12.83		  & \textbf{5.43}	& \textbf{66.9}& 8.33   & \textbf{8.14}	\\ \hline
\end{tabular}
}
\label{tab:debug_react}
}
\vspace{-10pt}
\caption[]{{
\footnotesize
\textbf{\subref{tab:best_strategies}} Percentage rates 
in how many debugging sessions 
which strategy performed best/better w.r.t.~the required user interaction, i.e. number of queries. EXP-1 and EXP-2 involved 27, EXP-3 and EXP-4 seven debugging sessions each. $q_{str}$ denotes the number of queries needed by strategy $str$ and $\min$ is an abbreviation for $\min(q_\text{SPL},q_\text{ENT})$.
\textbf{\subref{tab:debug_react}}
Average time (sec) for the entire debugging session (\textit{debug}), average time (sec) between two successive queries (\textit{react}), and average number of queries ($q$) required by each strategy.}}
\label{fig:best_diff}
\end{figure*}
\begin{figure*}[ht]
\vspace{7pt}
\centering
\subfigure[]{
    \includegraphics[width=0.33\textwidth,trim=7mm -12mm 1mm 20mm]{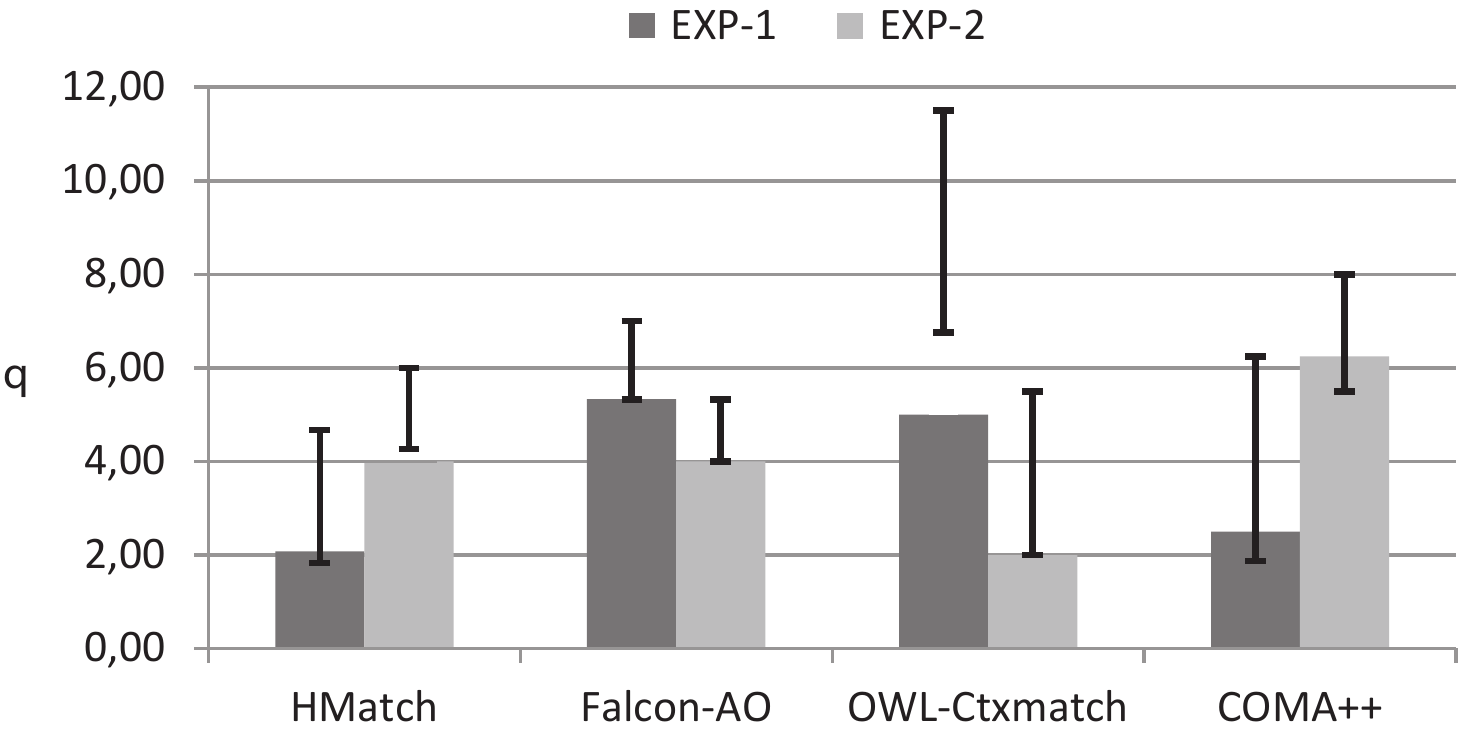}
		\label{fig:query_conference}
}
\subfigure[]{
    \includegraphics[width=0.33\textwidth,trim=5mm 7mm 1mm 13mm]{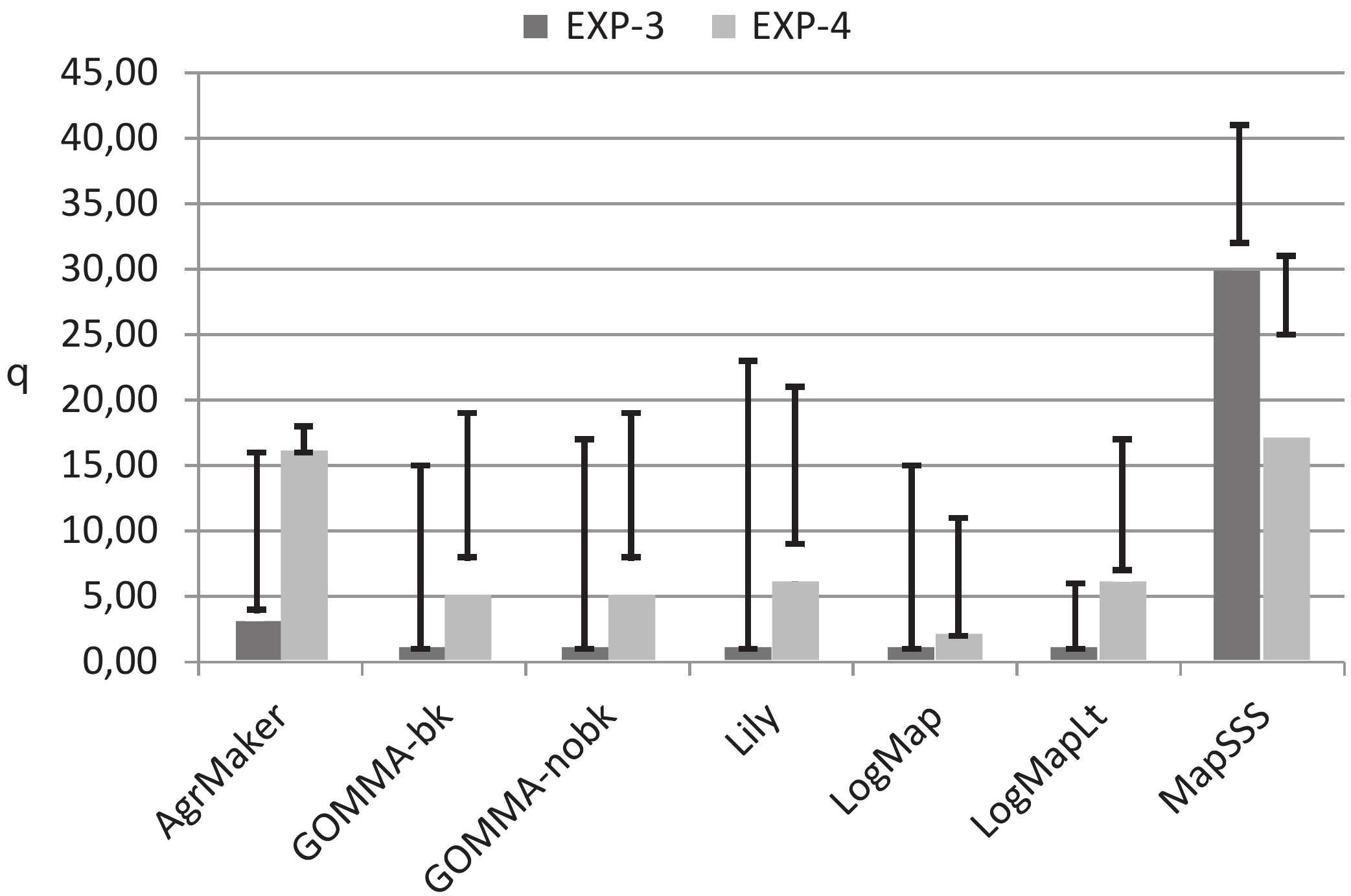}
    \label{fig:query_anatomy}
}
\subfigure[]{
    \includegraphics[width=0.30\textwidth, trim=2mm -11mm 0mm 15mm]{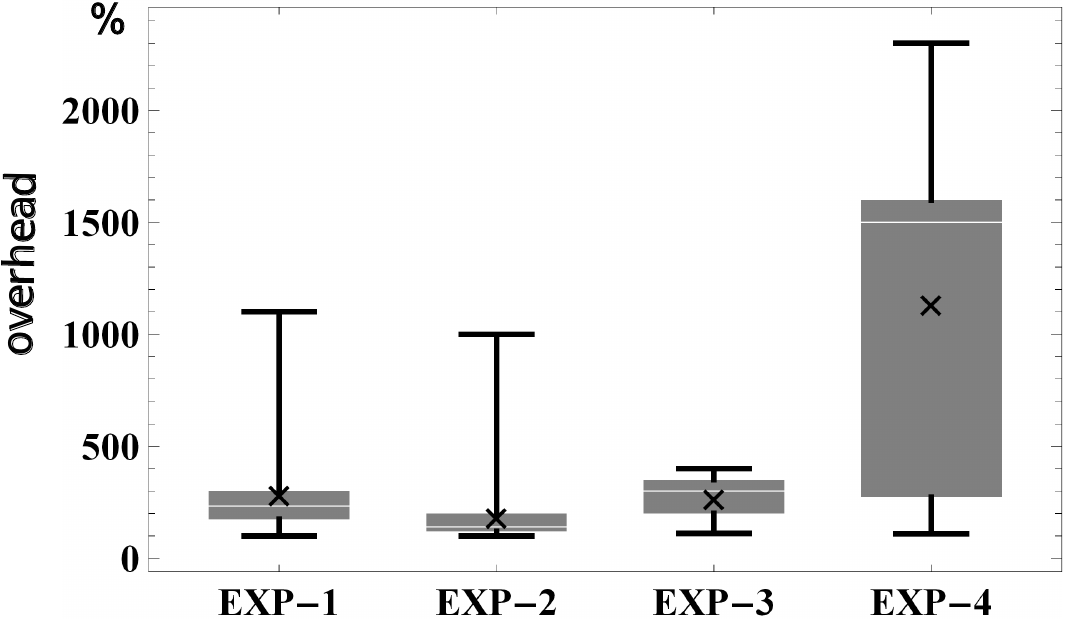}
    \label{fig:diff_SPL_ENT}
}
\vspace{-15pt}
\caption[]{{
\footnotesize
\textbf{\subref{fig:query_conference},\subref{fig:query_anatomy}} The bars show the average number of queries ($q$) needed by RIO, grouped by matching tools. 
The lower (upper) end of the whisker indicates the average $q$ needed by the per-session better (worse) strategy in \{SPL,ENT\}.
\textbf{\subref{fig:diff_SPL_ENT}} Box-Whisker Plots presenting the distribution of overhead $(q_w-q_b)/q_b*100$ (in \%) per debugging session of the worse strategy $q_w := \max(q_\text{SPL},q_\text{ENT})$ compared to the better strategy $q_b := \min(q_\text{SPL},q_\text{ENT})$. Mean values are depicted by a cross.}}
\label{fig:queries}
\vspace{-10pt}
\end{figure*}

\vspace{0pt}
\noindent\textbf{Test settings.}\footnote{See http://code.google.com/p/rmbd/wiki/ for code and details.}  We conducted four experiments EXP-$i$ ($i=1,\dots,4$), the first two with dataset D1 and the other two with D2. In experiments 1 and 3 we simulated good fault probabilities by setting $p(\tax_k) := 0.001$ for $\tax_k \in \mo_i \cup \mo_j$ and $p(\tax_m):= 1-v_m$ for $\tax_m \in \Al_{ij}$, where $v_m$ is the confidence of the correspondence underlying $\tax_m$. Low quality fault information was used in experiments 2 and 4. In \mbox{EXP-4} the following probabilities were defined: $p(\tax_k) := 0.01$ for $\tax_k \in \mo_i \cup \mo_j$ and $p(\tax_m):= 0.001$ for $\tax_m \in \Al_{ij}$. In EXP-2 we used probability settings of EXP-1, but fixed a completely unlikely target diagnosis in that we precomputed (prior to the actual experiment) the 30 most probable minimal diagnoses, and from these selected the one including the highest number of axioms $\tax_k \in \mo_{i\Al j} \setminus \N(\Al_{ij})$ as $\dt$. 

In all experiments, we set $|\mD|:=9$ which proved to be a good trade-off between computation effort and representativeness of leading diagnoses, $\sigma~:=~85\%$ and as input parameters for RIO $\uc:= 0.25$ and $[\underline{c},\overline{c}]:=[\underline{\qc},\overline{\qc}] = [0,\frac{4}{9}]$. To let tests pose the highest challenge for the evaluated methods, the initial DPI was 
specified~as $\tuple{\mo_{i \Al j},\emptyset,\emptyset,\emptyset}_{\setof{\text{coherence}}}$, i.e. the full search space was explored without adding parts of $\mo_{i \Al j}$ to 
$\mb$.
In practice, given prior knowledge of correct axioms, 
adding those to $\mb$ can severely restrict the search space and greatly accelerate debugging.
%
All tests were executed on a Core-i7 (3930K), 32GB RAM with Ubuntu 11.04 and Java~6.  

\noindent\textbf{Metrics.} Each experiment involved a debugging session of ENT, SPL as well as RIO for each ontology in the respective dataset. In each session we measured the number of required queries ($q$) until $\dt$ was identified, the overall debugging time (\textit{debug}) assuming that queries are answered instantaneously and the reaction time (\textit{react}), i.e. the average time between two successive queries. The queries generated in the tests were answered by an automatic oracle by means of the target ontology $\otarget:=\mo_{i\Al j}\setminus \dt$.

\noindent\textbf{Observations.} 
The difference w.r.t.~number of queries per test run between the better and the worse strategy in \{SPL,ENT\} was absolutely significant, with a maximum of 2300\% in EXP-4 and averages of 190\% to 1145\% throughout all experiments
(Figure~\ref{fig:queries}\subref{fig:diff_SPL_ENT}). 
Moreover, results show that varying quality of fault probabilities in \{\text{EXP-1},\text{EXP-3}\} compared to \{\text{EXP-2},\text{EXP-4}\} clearly affected the performance of ENT and SPL (see first two rows in Figure~\ref{tab:best_strategies}). 
This perfectly motivates why a risk-optimizing strategy is suitable.

Results of both experimental sessions, $\langle\text{EXP-1,EXP-2}\rangle$ and $\langle\text{EXP-3,EXP-4}\rangle$, are summarized in Figures~\ref{fig:query_conference} and \ref{fig:query_anatomy}, respectively. 
The figures show the (average) number of queries asked by RIO and the 
(average) differences to the number of queries needed by the per-session better and worse strategy in \{SPL,ENT\}, respectively. 
The results illustrate clearly that the average performance achieved by RIO was always substantially closer to the better than to the worse strategy. 
In both EXP-1 and EXP-2, throughout 74\% of 27 debugging sessions, RIO worked as efficiently as the best strategy (Figure \ref{tab:best_strategies}).  
In 26\% of the cases in EXP-2, RIO even outperformed both other strategies; in these cases, RIO could save more than 20\% of user interaction on average compared to the best other strategy. In one scenario 
in EXP-1, it took ENT 31 and SPL 13 queries to finish, whereas RIO required only 6 queries, which amounts to an improvement of more than 80\% and 53\%, respectively. In $\langle\text{EXP-3,EXP-4}\rangle$, the savings achieved by RIO were even more substantial. RIO manifested superior behavior to both other strategies in 29\% and 71\% of cases, respectively. Not less remarkable, in 100\% of the tests in EXP-3 and EXP-4, RIO was at least as efficient as the best other strategy.
Recalling Figure~\ref{fig:diff_SPL_ENT}, this means that RIO can avoid query overheads of 2200\%.
%
Figure~\ref{tab:debug_react}, which provides 
average values for $q$, \textit{react} and \textit{debug} per strategy,
demonstrates that RIO is the best choice in all experiments w.r.t. $q$. Consequently, RIO is suitable for both good 
and poor meta information.
As to time aspects, RIO manifested good performance, too. Since 
times consumed 
in $\langle\text{EXP-1,EXP-2}\rangle$ are almost negligible, consider the more meaningful results obtained in $\langle\text{EXP-3,EXP-4}\rangle$. While the best reaction time in both experiments was achieved by SPL, we can clearly see that SPL was significantly inferior to both ENT and RIO concerning $q$
and \textit{debug}. RIO revealed the best debugging time in EXP-4, and 
needed only $2.2\%$ more time than 
the best strategy (ENT) in EXP-3. However, if we assume the user being capable of reading and answering a query in, e.g., 30 sec on average, which is already quite fast, then the overall time savings of RIO compared to ENT in EXP-3 would already account for $5\%$. Doing the same thought experiment for EXP-4, 
RIO would save $25\%$ (w.r.t. ENT) and $50\%$ (w.r.t. SPL) of debugging time on average.
All in all, the measured times confirm that RIO is well suited for \emph{interactive} debugging.

\vspace{-5pt}

\section{Conclusions}
\label{sec:conclusion}
We have shown problems of state-of-the-art interactive ontology debugging strategies w.r.t.\ the usage of unreliable meta information. To tackle this issue, we proposed a learning strategy which combines the benefits of existing approaches, i.e. high potential and low risk. Depending on the performance of the diagnosis discrimination actions, the trust in the a-priori information is adapted. Tested under various conditions, our algorithm 
revealed good scalability and reaction time as well as superior average performance to two common approaches in the field w.r.t. required user interaction.  

\bibliographystyle{named}
\bibliography{V-Know}
\end{document}